\documentclass{article}
\usepackage{spconf,amsmath,graphicx}

\usepackage{color}
\usepackage{amssymb}
\usepackage{float}
\usepackage{afterpage}
\usepackage{xcolor}

\usepackage{fancyhdr}
\fancypagestyle{firstpagefooter}{

  \fancyhf{}
    \fancyhead[L]{Under review as a conference paper at ICASSP 2024}
  
  \fancyfoot[L]{\small 
  * Hao Wu is the correspondence author.
  \\© 20XX IEEE. Personal use of this material is permitted. Permission from IEEE must be obtained for all other uses, in any current or future media, including reprinting/republishing this material for advertising or promotional purposes, creating new collective works, for resale or redistribution to servers or lists, or reuse of any copyrighted component of this work in other works
  }
}
\title{Multi-Scale and Multi-Modal Contrastive Learning Network for Biomedical Time Series}
\name{Hongbo Guo\textsuperscript{1,2,3}, Xinzi Xu\textsuperscript{1,2}, Hao Wu\textsuperscript{3,4,*},  Guoxing Wang\textsuperscript{1,2,3}}
\address{\textsuperscript{1}School of Electronic Information and Electrical Engineering, Shanghai Jiao Tong University, Shanghai, China\\
\textsuperscript{2}Key Laboratory for Thin Film and Microfabrication of Ministry of Education,\\ Shanghai Jiao Tong University, Shanghai, China\\
\textsuperscript{3}Guangdong JiuZhi Technology Co., Ltd, Zhongshan, Guangdong, China\\
\textsuperscript{4}College of Electronics and Information Engineering, Shenzhen University, Shenzhen, Guangdong, China
}

\begin{document}
\thispagestyle{firstpagefooter}
\ninept
\maketitle
\begin{abstract}
Multi-modal biomedical time series (MBTS) data offers a holistic view of the physiological state, holding significant importance in various bio-medical applications. Owing to inherent noise and distribution gaps across different modalities, MBTS can be complex to model. Various deep learning models have been developed to learn representations of MBTS but still fall short in robustness due to the ignorance of modal-to-modal variations. This paper presents a multi-scale and multi-modal biomedical time series representation learning (MBSL) network with contrastive learning to migrate these variations. Firstly, MBTS is grouped based on inter-modal distances, then each group with minimum intra-modal variations can be effectively modeled by individual encoders. Besides, to enhance the multi-scale feature extraction (encoder), various patch lengths and mask ratios are designed to generate tokens with semantic information at different scales and diverse contextual perspectives respectively. Finally, cross-modal contrastive learning is proposed to maximize consistency among inter-modal groups, maintaining useful information and eliminating noises. Experiments against four bio-medical applications show that MBSL outperforms state-of-the-art models by 33.9\% mean average errors (MAE) in respiration rate, by 13.8\%  MAE in exercise heart rate, by 1.41\% accuracy in human activity recognition, and by 1.14\% F1-score in obstructive sleep apnea-hypopnea syndrome.

\end{abstract}
\begin{keywords}
bio-medical time series, multi-modal, representation learning, contrastive learning
\end{keywords}
\section{Introduction}
\label{sec:intro}
Multi-modal biomedical time series (MBTS) data have been widely used for various bio-medical applications, such as combining photoplethysmogram (PPG) and blood oxygen level (SpO$_2$)\cite{9964431} to jointly predict sleep apnea-hypopnea syndrome (OSAHS). 

Deep learning (DL) models especially temporal convolution networks (TCN)\cite{DBLP:journals/corr/abs-1803-01271}\cite{yue2022ts2vec} are widely used for MBTS data mining due to their good performance and low resource consumption, which uses dilated causal convolutions to capture long-term temporal dependency. To better capture extract multi-scale patterns (MS) in MBTS, MS extraction methods\cite{cui2016multiscale}\cite{Ismail_Fawaz_2020}\cite{osathitporn2023rrwavenet} are designed. Besides, to learn useful representations from MBTS, contrastive learning\cite{yue2022ts2vec}\cite{huang2021concad}\cite{zhang2022selfsupervised} is also often used.
%

%
However, the performance of existing models is limited due to the ignorance of the distribution gap across modalities and underfitted MS feature extraction modules.
Hence, to further improve the performance, this paper presents a multi-scale and multi-modal biomedical time series contrastive learning (MBSL) network which includes the following main parts. 1) \textbf{Inter-modal grouping technique to address distribution gap across modalities}. The amplitude range of MBTS, which may vary significantly across modalities, is unbounded. For example, SpO$_2$ values usually vary from 70\% to 100\%, while the acceleration values vary from 0 to 1000 $cm/s^{2}$. MBTS also features diverse structured patterns, such as the seasonally dominant PPG and trend dominant SpO2, suited for frequency and time domain modeling respectively\cite{2022Season}. Therefore, the commonly used parameter-sharing encoder for all modalities\cite{yue2022ts2vec}\cite{zhang2022selfsupervised} may limit the capability of the encoders to represent different modalities. An intuitive approach is to customize a specific encoder for each modality to extract heterogeneous features, but this approach would be inefficient and lack robustness\cite{han2023capacity}. To achieve a more capable and robust model, inter-modal grouping (IMG) is proposed to group MBTS into different groups with minimum intra-model distances, thus data in the same group manifest similar distributions, which is simple to represent with a shared encoder. Then distinct encoders are trained for each group to handle multi-modal data distribution gaps. 2) \textbf{A lightweight multi-scale data transformation using multi-scale patching and multi-scale masking.} Most MS extraction algorithms primarily rely on multi-branch convolution utilizing receptive fields of varying scales\cite{Ismail_Fawaz_2020}\cite{osathitporn2023rrwavenet}. However, it quadratically increases the memory and computing requirements. Some works like MCNN\cite{cui2016multiscale} use down-sampling to generate multi-scale data, but it would lose considerable useful information. This paper proposes a lightweight multi-scale data transformation using multi-scale patching and multi-scale masking, where different patch lengths are defined to obtain input tokens
with semantically meaningful information at different resolutions, and
the different mask ratios are adaptively defined according to the scale of features to be extracted. 3) \textbf{Cross-modal contrastive learning.} 
Inappropriate data augmentation can alter the intrinsic properties of MBTS, leading to the formation of false positive pairs. This issue causes many time series contrastive learning models to encode invariances that might not align with the requirements of downstream tasks.\cite{yang2022unsupervised}
For instance, excessive masking risks obscuring physiologically informative sub-segments.
To construct more reasonable contrastive pairs, cross-modal contrastive learning is utilized to learn modal-invariant representations that capture information shared among modalities. The intuition behind the idea is that different modalities commonly reflect the physiological state so the modal-invariant semantic is useful information for BMA.
\afterpage{
\begin{figure*}[ht]
\centering
  \includegraphics[width=0.7\textwidth]{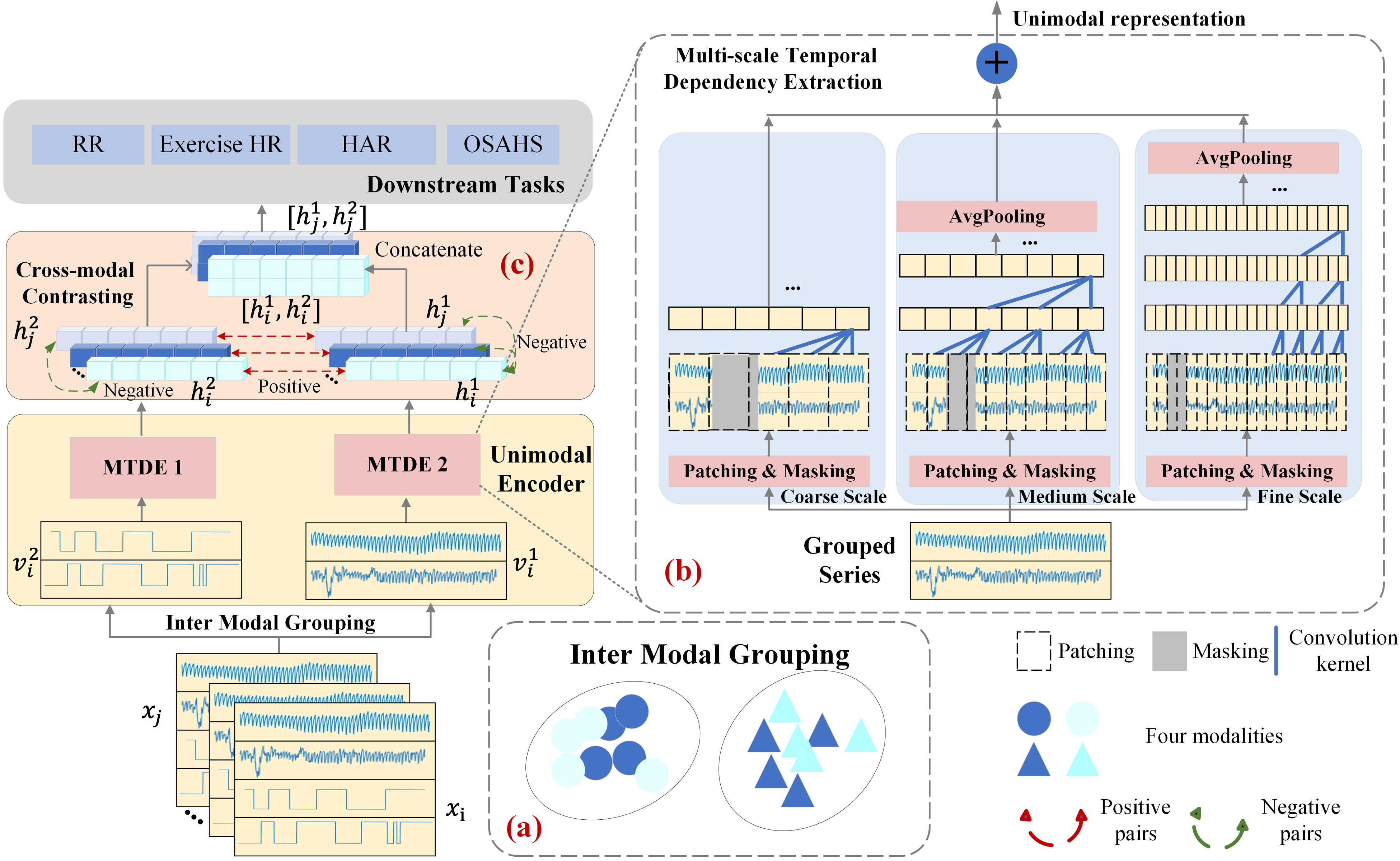}
  \caption{MBSL architecture. (a) Inter-modal grouping; (b) Multi-scale Temporal Dependency Extraction; (c) Cross-modal Contrasting. $x_{i}$ represents the i-th input series, $v_{i}^{1}$, and $h_{i}^{1}$ represents the raw data and embedding of the first group of the i-th sample, respectively. }
  \label{fig:mylabel}
\end{figure*}
}
The contributions are summarized as follows:
\begin{enumerate}
\item We propose multi-scale patching and multi-scale masking for extracting features at various resolutions and adaptively enhancing the capacity of the model, respectively.
\item We propose inter-modal grouping to process multi-modal data and use cross-modal contrastive learning for effective positive pairs construction and multi-modal data fusion.
\item Our model outperforms state-of-the-art (SOTA) methods across four datasets.
\end {enumerate}


\section{Method}
The overall MBSL architecture, illustrated in Fig. 1, includes IMG, multi-scale temporal dependency extraction (MTDE), and cross-modal contrastive loss, briefly described below.

\subsection{Inter-modal Grouping}
As mentioned before, the distribution gap severely impairs the generalization ability of the network. IMG is utilized to address this challenge. Intuitively, MBTS is divided into distinct grouped modalities based on their inter-modal distances. Modalities within each group exhibit similar distributions. Independent encoders are used to model different grouped modalities, respectively. Our approach aligns with the intuition that similar data should be processed similarly.

In detail, we consider a set of M modalities of the data, represented as $X_{1}$, . . . , $X_{M}$. Each modality is first dimensionally reduced using t-SNE, and then the inter-modal distance is calculated through the Euclidean distance. K group modalities $v_{1}$, . . . , $v_{K}$ will be obtained. Within a group, the minimum distance between any modality $X_{i}$ and other modalities $X_{j}$ ($i \neq j$ ) needs to be less than a threshold I:
\begin{eqnarray}
D_{i} = argmin\left\{d(X_{i},X_{j}\right\}_{j=1}^g < I
\end{eqnarray}
where $g$ is the number of modalities within a group. K group-specific encoders will be used for the K-grouped modalities.

\subsection{Improved multi-scale Temporal Dependency Extraction}
Recently, TCN has been proven to be a highly effective network architecture for BMA.
However, real-world MBTS is very complex, a unified receptive field (RF) within each layer of TCN is often not powerful enough to capture the intricate temporal dynamics. What’s more, the effective RFs of the input layers are limited, causing temporal relation loss during feature extraction. To adaptively learn a more comprehensive representation, we designed an MTDE module, which uses MS data transformation (patching and masking) and MS feature extraction encoder.

\subsubsection{Multi-scale patching and Multi-scale Masking}
\textbf{Multi-scale patching}  Different lengths of patches are used to transform the time series from independent points to tokens with semantic information within adaptive subseries, allowing the model to capture features of different scales. In particular, by aggregating more samples, long-term dependence with less bias can be obtained instead of short-term impact. Simultaneously, patching can efficiently alleviate the computational burden caused by the explosive increase in multiple TCN branches by reducing the sequence length from $L$ to $\frac{L}{P}$, where $P$ is the patch length.

\textbf{Multi-scale Masking} Masking random timestamps helps improve the robustness of learned representations by forcing each timestamp to reconstruct itself in distinct contexts, which can be formulated as $x_{{mask}_i}=\ x_i\ast m$, where $m$ $\in0,1$. 
Commonly, the ratio of masked samples to the whole series, denoted as $M_R$ is set to be small to avoid distortion of original data. However, as to different scales of features, the optimized $M_R$ is adaptive based on the experimental results. For instance, for coarser-grained features, larger $M_R$ is preferred so that ample semantic information can still be retained while introducing strong variances, thereby enhancing the robustness. Hence, different $M_R$s are selected to adapt to multi-scale features. 

\subsubsection{Multi-scale feature extraction encoder}
After MS data transformation, sketches of a time series at different scales are generated. Then MS time series pass through parallel TCN encoders with different kernel sizes and layers and then concatenated at the end. Average pooling is used to align sequence lengths to uniform lengths.

\subsection{Cross-modal Contrastive Loss}
Previous research on contrastive representation learning mainly tackled the problem of faulty positive pairs and noise in data. In the process of generating positive pairs by data augmentation, incorrect positive samples are also introduced, leading to suboptimal performance. 

In this work, the same segments from different modalities are chosen as positive pairs instead of generating them by data augmentation, thus avoiding the risk of introducing faulty positive pairs. Besides, this is beneficial to learn the semantic information shared by modalities because signals from different modalities of the same segments can intuitively reflect the subject's physiological state. Hence, during reducing the loss of contrastive learning, effective information among different modalities of the same segment will be maximized while redundant information across different segments will be ignored.

In particular, we consider all grouped modality pairs (i, j), $i \neq j$, and optimize the sum of a set of pair-wise contrastive losses:
\begin{eqnarray}
\mathcal{L}_{CM}=\sum_{0<i<j\leq_M}{\mathcal{L}_{V_{i},V_{j}}}
\end{eqnarray}

\begin{equation}
\mathcal{L}_{V_{1},V_{2}} = -\log \frac{\exp \left(\operatorname{sim}\left(h_{i}^{1}, h_{i}^{2}\right) / \tau\right)}{\sum_{k=1}^{N} \mathbb{1}_{\substack{k \neq i}} \exp \left(\operatorname{sim}\left(h_{i}, h_{k}\right) / \tau\right)}
\end{equation}

where $h_{i}^{1}$ is the embedding of the first group of the i-th sample, $\operatorname{sim}\left(h_{i}^{1}, h_{i}^{2}\right)=h_{i}^{1} h_{i}^{2} /\left\|h_{i}^{1}\right\|\left\|{h}_{i}^{2}\right\|$, $\mathbb{1}_{k \neq i} \ $in 0,1  is an indicator function evaluating to 1 if  $k \neq i$  and  $\tau$  denotes a temperature parameter, N represents the size of batch size.

\begin{figure}[!t]
  \includegraphics[width=\columnwidth]{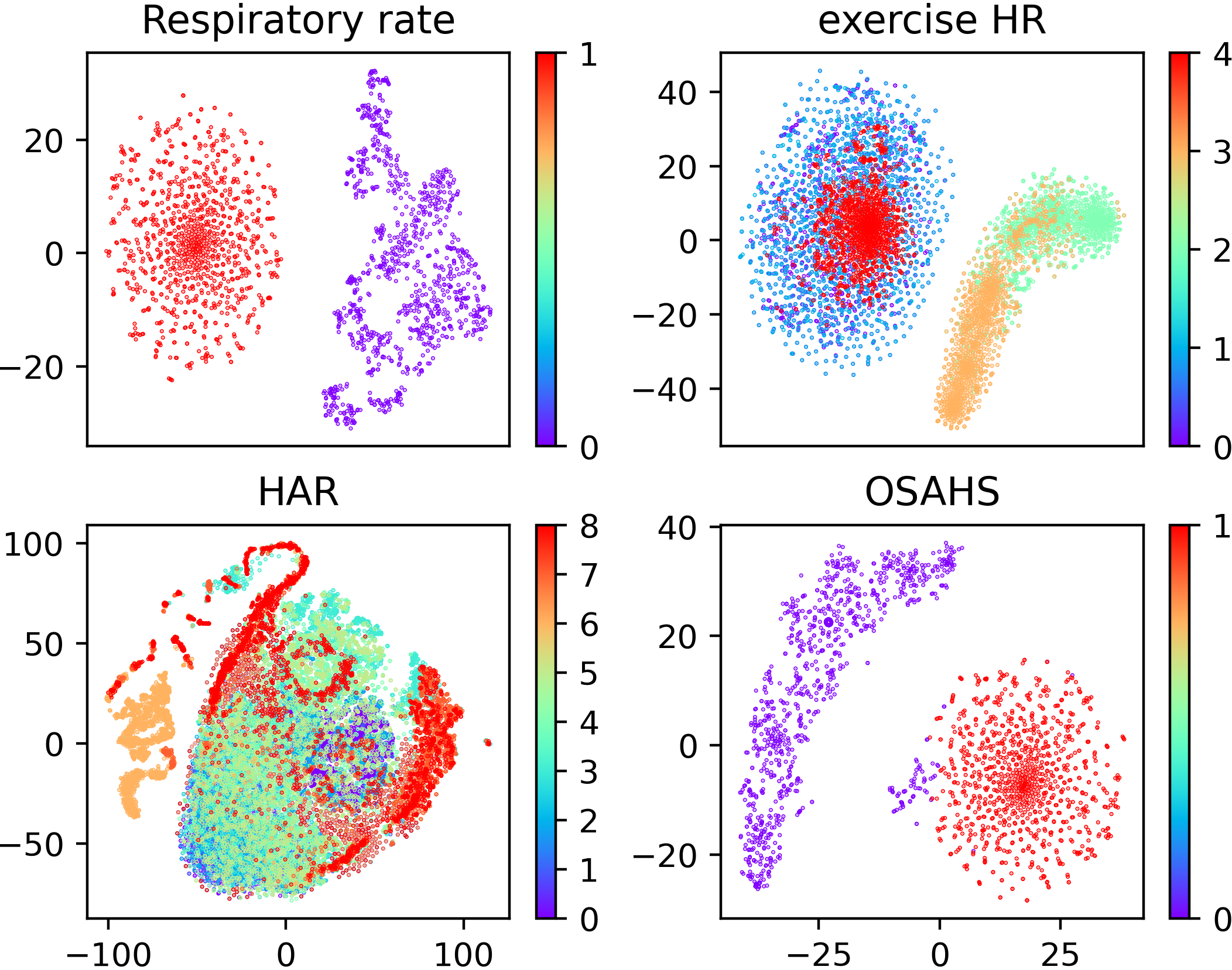}
  \caption{Distribution gap in multi-modal biomedical time series. Different colors represent different modalities in that dataset.}
  \vspace{-0.0cm}
\end{figure}

\section{Experimental Result}
\subsection{Experimental Setup} The size of the hidden layers and output layer is 32 and 64 respectively. The mask ratio, $M_R$ is 0.05, 0.1, and 0.15 and the patch length is $0.04*fs$, $0.08*fs$, and $0.16*fs$ for small, middle, and large scales, respectively, where $fs$ is the sample rate of the dataset. The temperature $\tau$ of the contrastive learning is 0.1.  The model is optimized using Adam with a batch size of 480 and a learning rate of 0.002. 

\subsection{Results}
The proposed MBSL is applied to MBTS datasets in two major practical tasks including regression and classification. The two latest contrastive learning works, i.e. TFC\cite{zhang2022selfsupervised}and TS2Vec\cite{yue2022ts2vec}, and one latest non-contrastive self-supervised learning work TimeMAE\cite{cheng2023timemae} are used as baseline models, and we also compare them with SOTA under the corresponding datasets. The t-SNE visualization of these datasets is shown in Fig. 2, which shows that there is a significant distribution gap between different modalities.

\subsubsection{MBTS regression} 
\begin{table}[!t]
\begin{center}
\vspace{0.0cm}

\caption{Results of RR detection. W represents the length of the signal. The best results are in \textbf{bold} and the second best are \underline{underlined}.}
\begin{tabular}{cccc}

\hline
\cline{1-4} 
Methods       & MAE (W=16)              & MAE (W=32)               & MAE (W=64)        \\  \hline     
TFC\cite{zhang2022selfsupervised}           & 4.16 $\pm$ 3.31          & 2.96 $\pm$ 3.19          & 4.54 $\pm$ 3.65    \\                  
TS2Vec\cite{yue2022ts2vec}        & 3.27 $\pm$3.35           & 2.56 $\pm$ 2.93          & 2.34 $\pm$ 3.01    \\              
TimeMAE\cite{cheng2023timemae}       & 1.82 $\pm$ 1.22          &   \underline{1.60 $\pm$ 1.49 }     & 2.02 $\pm$ 1.97                    \\
RespNet\cite{8856301}       & 2.45 $\pm$ 0.69          & 2.07 $\pm$ 0.98          & 2.06 $\pm$ 1.25  \\                
RRWAVE\cite{osathitporn2023rrwavenet}        & \underline{1.80 $\pm$ 0.95}          & 1.62 $\pm$ 0.86          & \underline{1.66 $\pm$ 1.01}   \\ 
\textbf{Ours} & \textbf{1.28 $\pm$ 0.84} & \textbf{1.13 $\pm$ 0.95} & \textbf{0.91 $\pm$ 0.81}                        \\ \hline
\end{tabular}
\vspace{0.0cm}
\end{center}
\end{table}

\begin{table}[!t]
\begin{center}
\caption{Results of exercise HR detection. The best results are in \textbf{bold} and the second best are \underline{underlined}.}
\vspace{0.3cm}
\begin{tabular}{cc}
\hline
 \cline{1-2} 
   Methods       & RMSE \\     \hline
   TFC\cite{zhang2022selfsupervised}           & 31.1                      \\
   TS2Vec\cite{yue2022ts2vec}        & 31.2                      \\
   TimeMAE\cite{cheng2023timemae}       & 30.9 \\
   TST\cite{zerveas2020transformerbased}           & 25.0                      \\
   InceptionTime\cite{Ismail_Fawaz_2020}     & \underline{23.9}                      \\
   \textbf{Ours} & \textbf{20.6}             \\ \hline
\end{tabular}
\vspace{-0.3cm}
\end{center}
\end{table}
We evaluate MBSL on open-source time series regression datasets: Respiratory rate (RR) detection\cite{7748483}, providing 53  8-minute PPG data segment (125 HZ) recordings, and exercise heart rate (HR) detection\cite{6905737}, including 3196 2-channel PPG and 3-axis acceleration sampled at a frequency of 125Hz.

\textbf{Evaluation method:} \textbf{1) RR} The PPG signal and its Fourier transform in frequency domain signal are used as multivariate biological time series. Leave-one-out cross-validation is applied and the preprocessing method follows \cite{osathitporn2023rrwavenet}. The average mean absolute error (MAE) and the standard deviation (SD) across all patients are used to evaluate the performance of the model \cite{osathitporn2023rrwavenet}.

\textbf{2) Exercise HR.} It is divided into two groups, one of which is the two-way PPG signal and one-way acceleration signal, and the other is the two-way acceleration signal.  We use root mean squared error (RMSE) to evaluate model performance and divide the dataset into the training set, validation set, and test set as in \cite{zerveas2020transformerbased}.

\textbf{Compare with the SOTA.}
Table 1 and Table 2 quantitatively demonstrate our method's superiority, reducing MAE in RR by 33\% and RMSE in exercise HR by 13.8\%.

\textbf{Case study.}
Exercise HR detection poses a challenge because physiological signals distort during exercise. As depicted in the first two rows of Fig. 3(a), recovering heart rate requires managing these altered signals. The acceleration signal plays a crucial role in this process. This scenario rigorously evaluates the model's proficiency in multi-scale feature extraction and multi-modal data processing. In Fig. 3(b), the regression results of TFC\cite{zhang2022selfsupervised} and TS2Vec\cite{yue2022ts2vec} on the test dataset appear as straight lines. InceptionTime\cite{Ismail_Fawaz_2020}, with its MS extraction model, delivers improved results. Owing to its enhanced multi-modal data processing and more robust MS feature extraction capabilities, MBSL exhibits the best performance.
\subsubsection{MBTS classification} 
We evaluate MBSL on widely-used MBTS classification datasets: HAR\cite{Anguita2013APD}, comprising 10,299 samples from 3-axis accelerometers, 3-axis gyroscopes, and 3-axis magnetometers, and OSAHS\cite{ross2019probabilistic} with 349,032 PPG and SpO2 records.

\textbf{Evaluation Method: 1) HAR.} It's split into two groups: one for a single axis of the 3-axis magnetometer and another for all other modalities. The dataset is divided as in \cite{eldele2021timeseries}. Accuracy (Acc) and the area under the precision-recall curve (AUPRC) are used to evaluate model like \cite{eldele2021timeseries}.
\textbf{2) OSAHS.}
It includes PPG and SpO$_2$ data, preprocessed as \cite{9492783} and split into four parts based on subject IDs—three for training/validation and one for testing. We use F1-score, recall (Re), and accuracy (Acc) to assess model performance.

\textbf{Compare with the SOTA.}
Table 3 shows that our MBSL improves accuracy by 1.88\% in HAR and increases recall by 2.13\%, F1-Score by 1.14\%, and accuracy by 0.46\% in OSAHS.
\begin{figure}[t]
\setlength{\abovecaptionskip}{0pt}
\setlength{\belowcaptionskip}{1pt}
  \includegraphics[width=0.9\columnwidth]{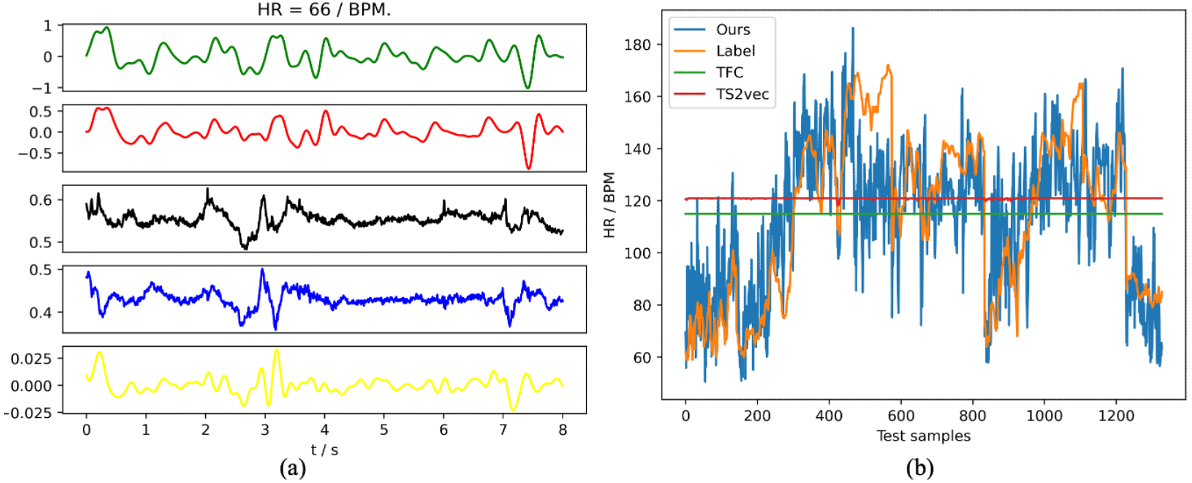}
  \caption{(a) A sample in the exercise heart rate data set. (b) Prediction results of different models on the whole exercise heart rate testset.}
\end{figure}

\begin{table}[t]
\caption{Result of HAR and OSAHS. The best results are in \textbf{bold} and the second best are \underline{underlined}.}
\tabcolsep=0.1cm
\vspace{0.3cm}
\begin{tabular}{ccc|cccc}

\hline
              & \multicolumn{2}{c|}{HAR}       & \multicolumn{4}{c}{OSAHS}                                        \\ \hline
Methods       & Acc            & AUPRC         & Methods       & F1             & Re             & Acc           \\
TFC\cite{zhang2022selfsupervised}           & 90.22          & 0.95          & TFC\cite{zhang2022selfsupervised}           & 62.30          & 57.76          & 82.38          \\
TS2Vec\cite{yue2022ts2vec}        & 90.57          & 0.85          & TS2Vec\cite{yue2022ts2vec}        & 60.57          & 55.40          & 81.82             \\
TimeMAE\cite{cheng2023timemae}       & \underline{95.10}          & \underline{0.99}          & TimeMAE\cite{cheng2023timemae}       & 74.41          & 75.43          & 86.01 \\
COT\cite{2023Cot}           & 94.05          & /          & ConCAD\cite{huang2021concad}        & 75.70          & \underline{76.54}          & 87.62          \\
BTSF\cite{yang2022unsupervised}           & 94.63          & \underline{0.99}          & U-Sleep\cite{9964431}       & \underline{76.20}          & 76.49      & \underline{87.92}          \\

\textbf{Ours} & \textbf{96.51} & \textbf{0.99} & \textbf{Ours} & \textbf{77.34} & \textbf{78.67} & \textbf{88.38} \\ \hline
\vspace{-0.3cm}

\end{tabular}
\end{table}

\begin{table}
\begin{center}

\caption{Ablation Experiments on Exercise Heart Rate Dataset. }
\vspace{0.3cm}

\begin{tabular}{ccc}
\hline
                                                                                    & RMSE   & MAE    \\ \hline
MBSL                                                                                & \textbf{20.6}  & \textbf{15.3}   \\ \hline
\multicolumn{3}{c}{\textbf{Inter-modal grouping}}                                                                              \\ 
w/o IMG                                                                      & 25.5  & 20.7   \\
Random Grouping                                                                        & 24.0  & 24.0   \\
Full Grouping                                                                             & 25.9  & 21.9   \\ \hline

\multicolumn{3}{c}{\textbf{Multi-scale temporal dependency extraction}}                                                                                    \\ 
w/o multi-scale masking                                                                            & 21.9 & 15.9  \\
\begin{tabular}[c]{@{}c@{}}Three moderate mask ratios\end{tabular}             & 21.3  & 16.0    \\
w/o multi-scale patching                                                                           & 24.1 & 18.5  \\
\begin{tabular}[c]{@{}c@{}}Three moderate patch lengths\end{tabular}          & 23.6  & 18.1    \\
MTDE—\textgreater{}TCN                                                              & 32.5  & 26.0   \\ \hline
\multicolumn{3}{c}{\textbf{Cross-modal contrastive loss}}                                                       \\ 
\begin{tabular}[c]{@{}c@{}}Supervised learning\end{tabular} & 22.2 & 16.3  \\ 
\begin{tabular}[c]{@{}c@{}}Cross-modal contrastive loss\\    
—\textgreater{} instance contrastive objective
\end{tabular}
& 22.4  & 16.2   \\ \hline

\vspace{-0.3cm}

\end{tabular}
\end{center}
\end{table}



\subsection{Ablation experiments}
Ablation experiments were conducted on the exercise HR dataset, given its complexity, to more effectively demonstrate the efficacy of our algorithm. We analyzed the impact of IMG, MTDE, and cross-modal contrastive loss, with results presented in Table 4. \textbf{IMG.} Excluding IMG significantly lowers performance due to distribution gaps. We also tested various grouping strategies: random grouping, where MBTS are arbitrarily grouped (matching IMG's group count), and full grouping, using five encoders for the five modalities in the exercise HR dataset. Both strategies showed declines, confirming that our grouped MI approach is effective and general. \textbf{MTDE.} Without patching and masking, MS feature extraction is limited, leading to poor performance. Using moderate mask ratios (0.1) and patch lengths (10) also reduces effectiveness, highlighting the need for varied data transformations for different scale feature extraction. Surprisingly, substituting MTDE with TCN results in a significant performance drop. \textbf{Cross-modal contrastive loss.} Eliminating or replacing this loss with standard instance contrastive loss leads to reduced model performance, underscoring the effectiveness of the cross-modal contrastive loss.

\section{Conclusion}
Deep learning is widely used in BMA. At present, the distribution gap across modalities and the complex dynamics of MBTS are still two major challenges for BMA. This article proposes a multi-scale and multi-modal representation learning network. In particular, the introduction of inter-modal grouping aims to bridge the distribution gap, resulting in a stronger and more robust model. Multi-scale patching and multi-scale masking are applied to extract features at various resolutions and avoid obscuring physiologically informative sub-segments while providing as strong as possible data augmentation, respectively. In addition, a cross-modal contrastive loss is used to encode the common semantic information across modalities, avoiding the risk of introducing faulty positive pairs by data augmentation. Evaluated on four datasets, our model outperforms current SOTA models, demonstrating the effectiveness of our approach.

\bibliographystyle{IEEEbib}
\bibliography{strings}

\end{document}